\documentclass{bmvc2k}
\usepackage{amsmath,amssymb,amsfonts}
\usepackage{algorithmic}
\usepackage{graphicx}
\usepackage{textcomp}
\usepackage{xcolor}
\usepackage{comment}
\usepackage{enumitem}
\usepackage{float}
\usepackage{hyperref}
\usepackage{placeins}
\usepackage{bmvc2k_natbib}
\usepackage[skip=0.15em]{caption}
\usepackage{stfloats}
\usepackage{float}

\usepackage{caption}
\usepackage{booktabs}
\title{Context-aware Bayesian Fusion for Multimodal Object Detection}

\addauthor{Arunachalam Venkatachalam}{Arunachalam.Venkatachalam@imec-int.com}{1}
\addauthor{Mhafuzul Islam}{mhafuzul.islam@imec-int.com}{1}
\addauthor{Marwan Yusuf}{marwan.yusuf@imec.be}{3}
\addauthor{David Van Hamme}{david.vanhamme@imec.be}{3}
\addauthor{Ben Stoffelen}{ben.stoffelen@imec.be}{2}
\addauthor{James Foresi}{Jim.Foresi@imec-int.com}{1}

\addinstitution{
 imec\\
 Michigan, USA
}
\addinstitution{
 imec\\
 Leuven, Belgium
}
\addinstitution{
 imec and Ghent University\\
 Ghent, Belgium
}

\runninghead{Venkatachalam et al.}{Context-aware Bayesian Fusion}

\begin{document}

\maketitle

\begin{abstract}
Reliable detection of pedestrians and road users is essential for the safe operation of autonomous vehicles. However, this task becomes increasingly challenging under adverse environmental conditions, such as poor weather and low lighting—where the performance of state-of-the-art pre-trained object detectors and on-board sensors tends to degrade, which limit their effectiveness and reliability. Moreover, retraining models for each operating context imposes significant constraints due to high data requirements, computational resources, and time. Considering this, we propose a novel context-aware multimodal late fusion approach that integrates a lightweight gating network within our probabilistic calibration framework. The gating network extracts environmental context and dynamically modulates the detection outputs of pre-trained object detectors based on the inferred context, without retraining of the detectors. Our methodology is evaluated on a modified KITTI dataset, in which the camera images and lidar pointclouds are augmented for varying fog levels. Experimental results show that our approach yields a ~14\% relative improvement over non-adaptive late fusion baselines, achieving a consistent average precision of $56\%$ across different fog levels. 
\end{abstract}

\section{Introduction}
\label{sec:intro}
In autonomous systems and intelligent sensing applications, robust object detection under adverse environmental conditions remains a critical challenge. Multimodal sensor fusion has emerged as a promising approach to enhance detection reliability by leveraging complementary information from heterogeneous sensor modalities. Sensor fusion can be achieved using early, mid and late fusion approaches, and according to recent studies late fusion is more accurate and cost-effective \cite{huang2022multimodal} \cite{wang2024review}. However, sensor performance can be highly dependent on environmental context, and sensor fusion approaches do not account for it in their pipelines. Some recent studies \cite{palladin2024samfusion} \cite{sural2024contextfusion} incorporate environmental context, but these deep neural network based methods require retraining to handle environmental conditions not present in the training data.  

Therefore, we propose a novel multimodal fusion framework that integrates pretrained object detectors operating on RGB cameras, and lidar, while dynamically adapting to environmental context.

\begin{figure}[ht]
    \centering
    \includegraphics[width=0.65\textwidth]{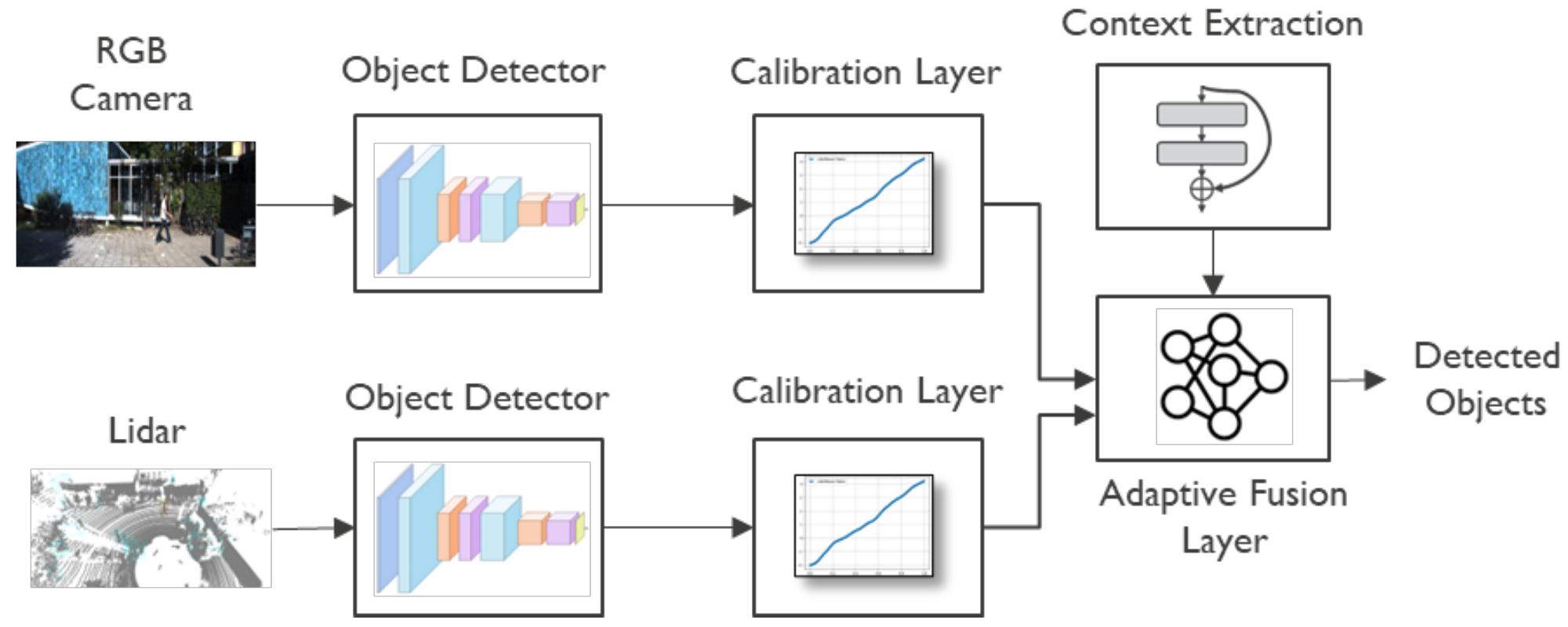}
    \caption{Overall system diagram for the adaptive multimodal sensor fusion.}
    \label{Figire_1}
\end{figure}

Our approach begins by independently applying pretrained object detectors to each sensor modality as shown in Figure \ref{Figire_1}. These detectors provide confidence scores for each detected object, which are then calibrated using a probabilistic transformation or calibration layer that converts raw confidence values into log-likelihood ratios. This calibration compensates for biases inherent in pretrained models and unifies the outputs across modalities. To further refine the fusion process, we introduce a neural-network based adaptive layer. This layer ingests both the LLRs, and contextual information, such as lighting conditions, weather, and visibility, which are extracted from the RGB cameras to dynamically adjust the contribution from each sensor. By learning to emphasize or de-emphasize sensor inputs based on situational awareness, the system achieves greater robustness and flexibility. Finally, a fusion layer aggregates the adaptively weighted outputs from all sensors to produce a unified detection result. This architecture enables context-aware, confidence-calibrated sensor fusion, improving object detection accuracy and reliability across diverse operational scenarios, specifically for adverse weather conditions. 

This paper makes the following key contributions:
\begin{enumerate}[noitemsep]
    \item We propose a context extraction network that captures fine-grained local context to better characterize varying sensing conditions. 
    \item We develop a mechanism to quantify how sensing modalities degrade with respect to this extracted context (e.g., lidar in fog).
    \item We introduce a method to dynamically calculate how to compensate for this in probabilistic fusion, enabling more robust cross-modality performance. 
\end{enumerate}

\section{Related Studies}

We classify prior research in multimodal object detection on the basis of operating environment (clear vs adverse weather) and underlying estimation methodology (deterministic vs probabilistic). Specifically, we focus on (i) standard fusion architectures for clear conditions, (ii) fusion architectures for adverse-weather, and (iii) fusion architectures with probabilistic methods for uncertainty estimation. Our work lies at the intersection of these threads, which we review below to position our contribution against the open gaps in the literature.

\subsection{Standard Multimodal Sensor Fusion}

Multimodal fusion has emerged as the dominant paradigm for safety-critical autonomous systems, leveraging the complementary strengths of cameras, lidars, and radars, with additional modalities such as gated NIR sensors being increasingly introduced \cite{huang2022multimodal} \cite{wang2024review}. Early and mid-fusion methods such as MV3D \cite{Chen_2017_CVPR}, AVOD \cite{ku2018joint}, F-PointNet \cite{fpointnet}, and transformer-based detectors (BEVFusion \cite{liang2022bevfusion}, CMT \cite{yan2023cross}, FusionFormer \cite{hu2023fusionformer}) achieve state-of-the-art accuracy on clear-weather benchmarks like KITTI and NuScenes by deeply coupling cross-modal features. However, this tight coupling makes them brittle to asymmetric sensor degradation. For instance, if one modality is corrupted, the joint feature space is compromised, often requiring the entire network to be retrained on data capturing the corrupted operating distribution. Conversely, late fusion approaches in \cite{chen2021multimodal} \cite{shaikh2022probabilistic} \cite{yang2025mmlf} decouple the modalities at the decision level, making them modular, computationally light, and natively compatible with off-the-shelf pretrained detectors — though it comes at the cost of discarding valuable low-level cross-modal cues.

\subsection{Probabilistic and Bayesian Late Fusion}


Closest to our framework are probabilistic late-fusion methods that combine outputs of independent detectors under Bayesian hypothesis. \cite{chen2022multimodal} formulates a non-learned Bayesian fusion rule derived from a conditional-independence assumption across modalities, reporting a performance gain of over 13\% in RGB-thermal pedestrian detection. \cite{shaikh2022probabilistic} establishes a naive Bayes decision-level fusion framework for color and thermal detectors that works without retraining with graceful handling of sensor failure.  \cite{willems2023bayesian} \cite{willems2024contextcompensation} develop a likelihood-ratio calibration layer that converts raw detector confidence scores into log-likelihood ratios (LLRs), which can be combined across sensors via Bayes' rule. Our proposed framework builds directly upon this approach.

\subsection{Adverse-Weather and Adaptive Fusion}

Recent research has shifted focus toward multimodal perception in degraded environments such as fog, rain, snow, and low light conditions. \cite{seeingthrufog} presents the Seeing Through Fog (STF) dataset and proposes an entropy-driven adaptive fusion that learns from clean data and generalizes to unseen weather. Another study, SAMFusion  \cite{palladin2024samfusion} extends this line with a transformer decoder that re-weights RGB, Lidar, gated-NIR, and radar features by distance and visibility, reporting +17.2 AP for distant pedestrians in fog. ContextualFusion \cite{sural2024contextfusion} injects coarse domain knowledge about modality reliability (e.g., Lidar degrades in fog, camera struggles under low-light) into a 3D detector and demonstrates gains on the simulated CARLA-Adverse and naturalistic NuScenes datasets. FogFusion \cite{zhang2026fogfusion} uses depth-completion with fog-aware convolutions to densify lidar before camera fusion, L4DR  \cite{huang2025l4dr} fuses Lidar with 4D radar for weather robustness, UAMF \cite{Li2025} introduces Bayesian uncertainty modeling with cross-modal attention conditioned on weather priors, and ACMD \cite{ghaffar2026adaptive}  performs adaptive cross-modal denoising guided by residual-noise estimation.

\subsection{Research Gap}

Existing adverse-weather and adaptive-fusion methods are powerful but require end-to-end retraining and rely on heuristic, scene-level weighting without probabilistic semantics. While existing Bayesian late-fusion methods preserve interpretability and reuse pretrained detectors, they assume context-invariant likelihoods and fail to adapt when sensing conditions shift. No prior work simultaneously offers: (i) Use of off-the-shelf pretrained single-sensor detectors with no fine-tuning, avoiding the high data and compute cost of retraining for new operating contexts, (ii) A context-dependent compensation factor derived directly from Bayes' rule rather than using a heuristic gating mechanism, providing the fused score with a well-defined posterior-ratio interpretation, and (iii) A context adaptation mechanism that extracts context from the local image region surrounding each candidate detection rather than a single scene-level weather descriptor (such as a global weather label), allowing the model to handle within-scene variations (e.g., a partially occluded pedestrian, a near vs. far target in patchy fog). To our knowledge, this is the first multimodal pedestrian-detection framework that combines pretrained-detector reuse, first-principles probabilistic compensation, and fine-grained local context extraction.

\begin{table}[ht]
\centering
\renewcommand{\arraystretch}{1.3} 
\resizebox{\textwidth}{!}{
\begin{tabular}{llll}   
\toprule
\textbf{Method Family} & \textbf{Pretrained Detectors?} & \textbf{Context-Aware?} & \textbf{Local Context?} \\
\midrule
Deep Mid-Fusion \scriptsize{(e.g., BEVFusion, MV3D)} & No \scriptsize{(requires retraining)} & No & No \\
Adverse-Weather Adaptive \scriptsize{(e.g., STF, SAMFusion)} & No \scriptsize{(requires retraining)} & Yes \scriptsize{(scene-level)} & No \\
Gating / Mixture-of-Experts \scriptsize{(e.g., Mees, GRIF)} & No \scriptsize{(requires retraining)} & Implicit & No \\
Bayesian Late Fusion \scriptsize{(e.g., Chen, Shaikh)} & Yes & No & No \\
\textbf{Proposed Method (Ours)} & \textbf{Yes} & \textbf{Yes} & \textbf{Yes} \\
\bottomrule
\end{tabular}
} 
\caption{A comparison of existing methods and our approach.}
\label{tab:method_comparison}
\end{table}

\section{Method}
In this section, we describe the theoretical formulation of the context-adaptive multimodal sensor fusion and implementation of our developed approach. 

\subsection{Bayesian Fusion}

Pre-trained object detectors perform differently under different scenarios, e.g., daytime vs nighttime performance, up-close vs long-range detection. To address this variability, the study in \cite{willems2023bayesian} introduced a calibration layer for single-sensor object detection by formulating a probabilistic decision framework based on Bayesian inference. This calibration layer enables translation of raw detection scores from pre-trained off-the-shelf object detectors into conditional probabilities of presence/absence of objects. By incorporating scenario information within a small subset of the data, the calibration layer adjusts the detector’s output to better reflect the true likelihood of presence of an object under varying environmental conditions.

\subsubsection{Theory}

Let $H_1$ denote the hypothesis that a pedestrian is present in a part of the scene, and $H_0$ the complementary hypothesis that no pedestrian is present. A pre-trained object detector produces candidate detections represented by tuples $y_l = (u_l, s_l, a_l)$ where $u_l$ is the location in sensor coordinates, $s_l$ is the size of an object in sensor coordinates, and $a_l$ is the confidence score in the range [0, 1]. These candidate detections can then be evaluated probabilistically against the hypotheses to assess the level of evidence for pedestrian presence they represent. With prior probability $P(H)$ representing the initial belief about presence or absence of a road user and marginal likelihood $p(y_l)$ representing the overall probability of obtaining an observation/detection under all hypotheses, the posterior probability of presence given the observation is expressed in terms of likelihood ratios and prior distributions using Bayes' rule as

\begin{equation}
P(H_1|y_l) =\frac{p(y_l|H_1) P(H_1)}{p(y_l)}
\end{equation}
and the posterior probability of absence with the same expression for $H_0$.

A detection should be considered positive if the following condition is satisfied: 

\begin{equation}
{P(H_1| y_l)} > P(H_0| y_l)
\end{equation}

From here onwards, we will focus only the detection confidence score $a_l$, omitting $u_l$, and $s_l$ since most object detectors are insensitive to location and trained to be insensitive to object size within a reasonable range. The decision boundary for candidate detections then becomes

\begin{equation}
\frac{p(a_l| H_1)} {p(a_l| H_0)} \lessgtr \frac{P(H_0)}{P(H_1)}.
\end{equation}

The functions $p(a_l| H_1)$ and $p(a_l| H_0)$ are called the likelihood functions and their ratio will be referred to as the \textit{likelihood ratio}.
Since the prior probabilities are independent from the input $a_l$ (they can broadly be interpreted as representing a sensitivity level for the decision), all information necessary for the calibration layer is contained in the likelihood ratio.

In this study, we extend this calibration framework to support multimodal fusion. Calibrated decision scores from each modality are combined to produce a unified estimate of the object presence. Assuming conditionally independent detections in the different modalities with confidence scores $a_1$ (i.e., lidar), $a_2$ (i.e., camera), the joint posterior probabilities are given by the product of the individual likelihood functions with the priors:


\begin{equation}
\frac{P(H_1 | a_1, a_2)}{P(H_0 | a_1, a_2)} = \frac{P(H_1)}{P(H_0)} \prod_{i=1}^{2} \frac{p(a_i | H_1)}{p(a_i | H_0)} 
\end{equation}

and the optimal decision criterion becomes
\begin{equation}
\prod_{i=1}^{2} \frac{p(a_i | H_1)}{p(a_i | H_0)} \lessgtr \frac{P(H_0)}{P(H_1)}.
\label{product_lr}
\end{equation}


\subsubsection{Implementation}\label{bayesian_fusion_impl}

To calculate the likelihood functions, a small number of negative (no person present) and positive (person present) samples of the ground truth data are evaluated by the object detector. For each set, a histogram of the detection scores produced for these samples is made and normalized into a density function. The likelihood ratio for any confidence value produced by the detector is then approximated by the ratio of the density function value of the positive set for this confidence to the density function value of the negative set for this confidence. This likelihood ratio function is constructed for each of the sensor modalities using a publicly available pre-trained object detector for each individual sensor.

In our implementation the likelihood ratios are converted to their logarithm to simplify computation and to aid numerical stability when potentially combining very small with very large values. We will hereafter refer to Log-Likelihood Ratios (LLRs).

\subsection{Context Extraction and Adaptation  }

\subsubsection{Theory}
It is observed that the distribution of confidence scores obtained can change significantly with respect to the sensing conditions, i.e., the scores depend on the \textit{observation context}. The posterior probabilities in our decision also depend on these conditions and the derivation should be modified to model the conditional dependence of the likelihood functions on the observation context. In principle, the context dependent likelihood ratios of every sensor can be obtained by splitting a sufficiently varied dataset into different context categories but in practice this strongly reduces the number of samples in every context category, leading to reduced representativeness of the histograms, and additionally requires one to decide the split into relevant contexts up front. Instead, we propose a compensation strategy that applies a modifier to the generic likelihood ratios of each sensor. Let the local, immediate context be described by a variable $c$. The context-dependent decision criterion is then given by
\begin{equation}
\frac{P(H_1|a_1, a_2, c)}{P(H_0|a_1, a_2, c)} 
= \frac{p(a_1, a_2, c|H_1) P(H_1)}{p(a_1, a_2, c|H_0) P(H_0)} 
\lessgtr 1
\end{equation}

The conditional independence assumption becomes stronger once context is introduced, as it is assumed to fully capture the sensors dependency. Hence assuming conditionally independent sensors, the likelihood ratio can be decomposed as

\begin{equation}
\frac{p(a_1, a_2, c|H_1)}{p(a_1, a_2, c|H_0)} 
= \frac{p(a_1,c |H_1)}{p(a_1,c|H_0)} \frac{p(a_2, c|H_1)}{p(a_2, c|H_0)} \frac{p(c)}{p(c)}.
\end{equation}


which in turn can be rewritten as

\begin{equation}
\left[ \frac{p(c|a_1, H_1)}{p(c|a_1, H_0)}  \frac{p(a_1|H_1)}{p(a_1|H_0)} \right] 
\left[ \frac{p(c|a_2, H_1)}{p(c|a_2, H_0)}  \frac{p(a_2|H_1)}{p(a_2|H_0)} \right].
\end{equation}

\begin{figure}[htbp]
\centerline{\includegraphics[width=0.65\columnwidth]{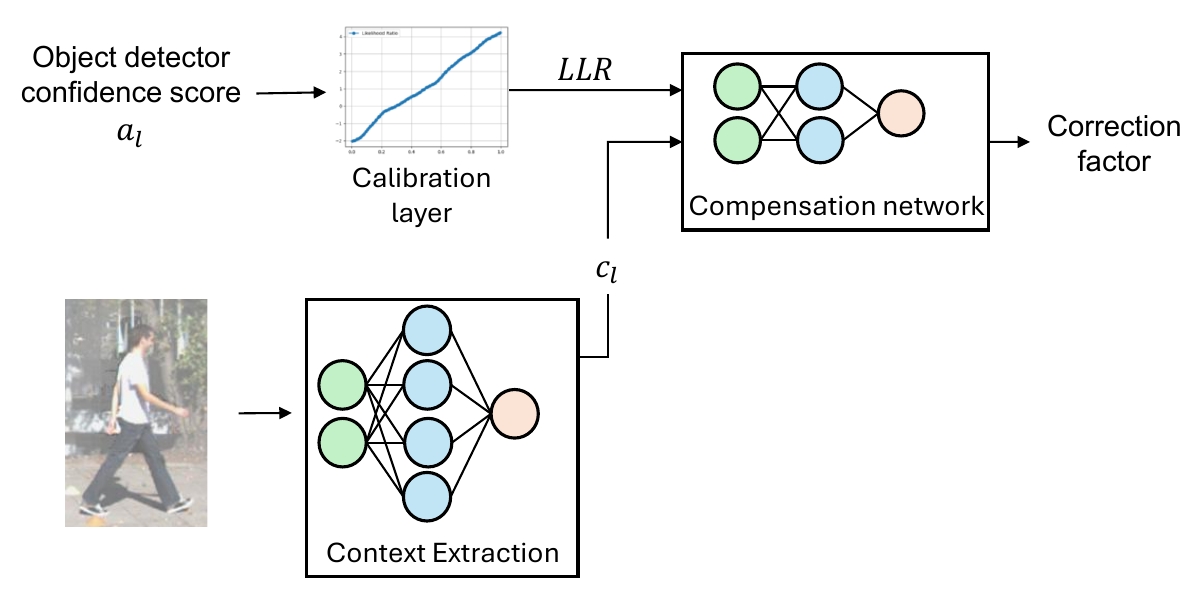}}
\caption{Overview of the context-aware probabilistic compensation network.}
\label{Fig_3_Overview}
\end{figure}

Comparing to equation \ref{product_lr}, we see that every sensor's likelihood ratio needs to be multiplied with a correction factor $p(c|a, H_1) / p(c|a, H_0)$. The principle of applying a multiplicative weight to each sensor's assessment is similar in structure to a mixture-of-experts ensemble model, where various expert models (in this case the sensors' detection algorithms) can have different levels of expertise for different types of input samples. Note that it is impractical to estimate the correction factors for all relevant contexts directly from labeled data because this suffers from the same histogram sparseness that prevents direct measurement of context-adaptive likelihood functions. Instead, we propose a two-tiered neural network structure to predict the factors (as shown in Fig \ref{Fig_3_Overview}):
\begin{enumerate}[noitemsep]
    \item a context extraction network that summarizes the local, immediate context pertaining to a detection in a continuous variable $c$,
    \item a compensation network that predicts for any $c$ the necessary correction factor for the not-context-specific likelihood function.
\end{enumerate}




This structure can approximate context-dependent likelihood ratios without explicit distribution modeling or model retraining, and adjusts the relative importance of each sensor in the Bayesian fusion framework.

Note that the relationship between the weight factor and the context variable $c$ can be highly non-linear and additionally depends on the likelihood ratio itself. The function of the weight can be seen as bringing the likelihood ratio closer to 1 (i.e equal probability of presence and absence) for contexts that reduce the reliability of the detector, and the required weight is thus different depending on whether the likelihood ratio is larger or smaller than one for the sample considered.

\subsubsection{Implementation}\label{context_impl}
\begin{figure}[htbp]
\centerline{\includegraphics[width=0.65\columnwidth]{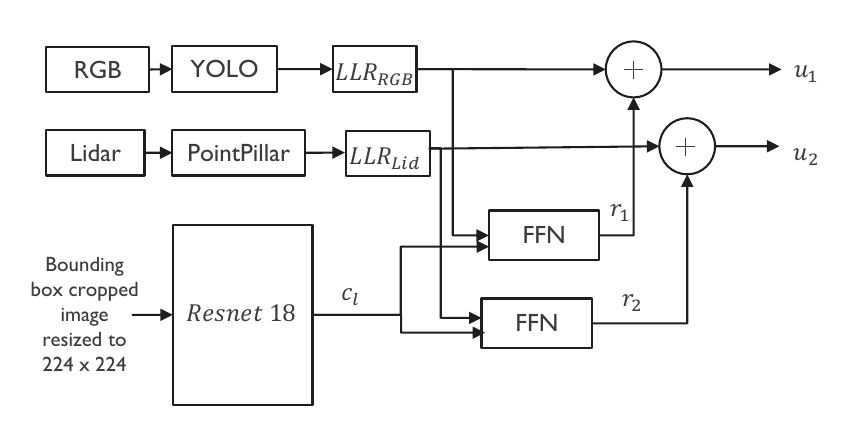}} 
\caption{Architecture of the implemented context-aware compensation network.}
\label{Fig_4_gating_net}
\end{figure}

Detection candidates across the sensor modalities are first matched and associated to produce candidate fused pairs. The region in the RGB image corresponding to each candidate pair is extracted from the RGB image and passed through the context extraction network which outputs a context value $c_l$ that characterizes the current scene context associated with each detection. This context extraction network is based on a ResNet18 architecture with the final layer modified to produce a single scalar. Note that since the context is extracted from the RGB crop, it may include object appearance, distance, occlusion, or detector artifacts — not only environmental context. Secondly, a custom feedforward network takes in the context value produced by the context extraction network along with the sensor-specific LLR values computed under clear weather conditions. The input vector is defined as $[c_l,LLR_m]$ where $m$ corresponds to the sensor modality (e.g. RGB, or lidar). To accommodate multiple sensor modalities, the architecture employs $n$ independent feedforward networks—one per modality—each acting as a separate adaptation head. These heads produce modality-specific weights $[r_1,r_2]$  that recalibrate the corresponding LLRs to align with the current environmental context. Note that implemented in the log-domain, the adaptation factors become bias terms. The complete architecture is shown in Figure \ref{Fig_4_gating_net}.

The context extraction network and the FFN compensation network are jointly trained, ensuring that the context variable $c_l$ captures only context information that is necessary to compensate the LLRs, without containing redundant information that is already assessed by the individual sensors' pre-trained detectors and thus represented in the original LLRs. To ensure that the compensation terms together with the original LLRs converge to fused LLRs, we use the loss function proposed in \cite{moustakides2019lossfn}. Note that in principle, the compensation can also be implemented as a single bias term $r$, but keeping separate terms for each sensor helps interpretability and aids convergence as the FFNs can be trained individually.





\section{Experimental Setup}
We evaluate the performance of the context-adaptive fusion framework for pedestrian detection in real-world scenarios using RGB image, and lidar pointcloud as input modalities. A key underlying assumption of this framework is that different modalities vary in how informative they are across contexts. By accordingly changing the relative impact each sensor has on the final decision, our method aims to reach better precision/recall tradeoffs across all conditions. 


Initial detection candidates and confidence scores are obtained using widely adopted pre-trained single sensor object detectors: YOLOv8m model \cite{redmon2016yolo} pre-trained on the COCO dataset for camera; and KITTI pre-trained PointPillars \cite{lang2019pp} for lidar. 

\subsection{Dataset}
We use the 3D object detection dataset from KITTI Vision Benchmark Suite \cite{geiger2012kitti} as it provides time-synchronized data from a RGB camera (Point Grey Flea 2) and a lidar (Velodyne HDL-64E ), accompanied by robust ground truth annotations and calibration matrices for each frame. The dataset consists of 7841 training images and 7518 test images. 


We subdivide KITTI training samples into three splits: training (60\%) with 4488 images, validation (20\%) with 1496 images, and testing (20\%) with 1497 images, as ground truth labels are required to evaluate performance.

\subsection{Fog Augmentation}

\begin{figure}[b]
\centerline{\includegraphics[width=0.75\columnwidth]{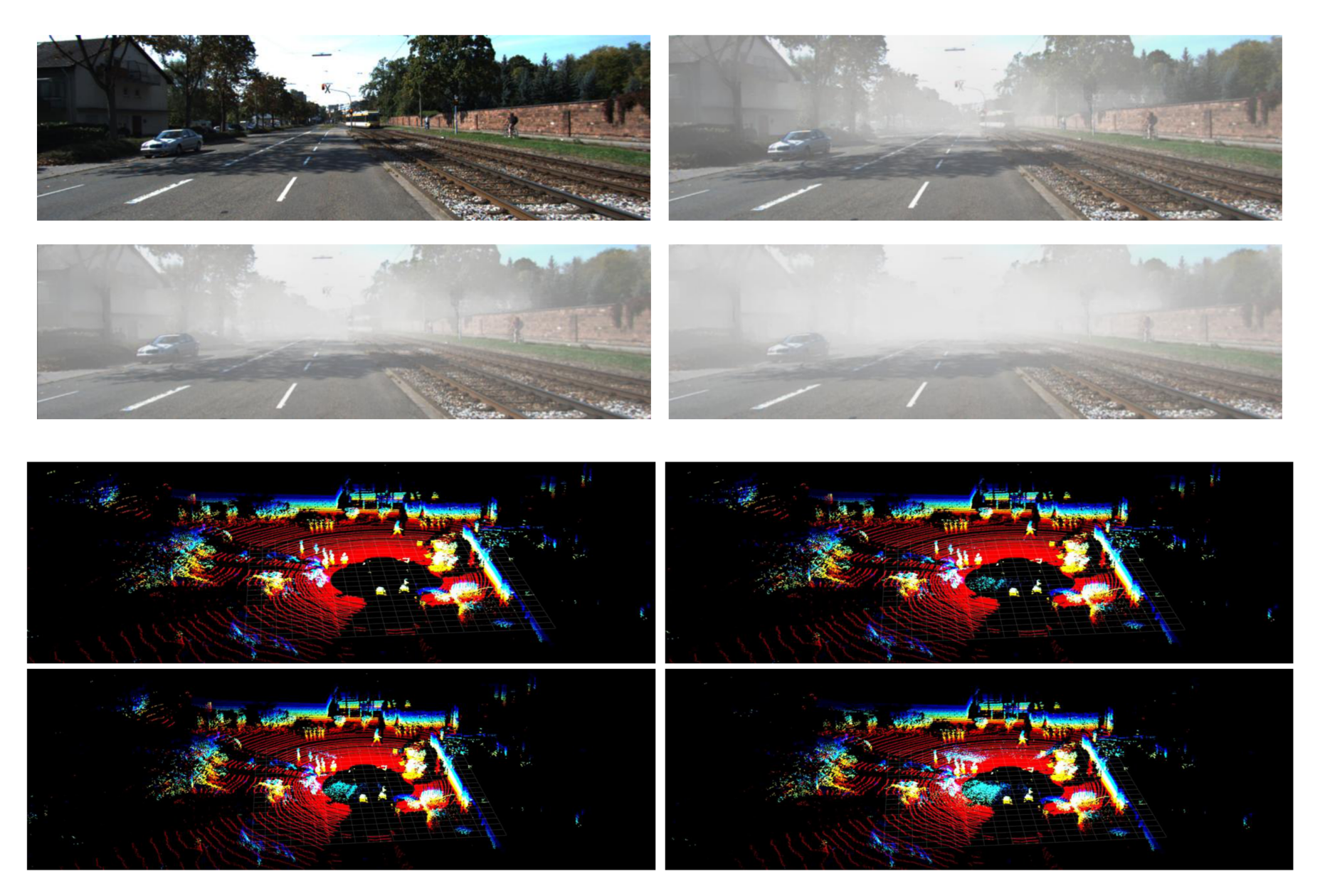}}
\caption{Sample images and pointclouds under varying visibility conditions. Top-left to bottom-right: clear, 75m, 50m, 30m.}
\label{Fig_6}
\end{figure}

Due to the challenging task of reliably labeling real-world fog datasets for 3D object detection, synthetic fog was generated by augmenting the original KITTI images and point clouds as shown in Figure \ref{Fig_6}. To characterize the severity of the fog, we use the commonly used metric visibility level \cite{wmo2024guide}, which quantifies the density of fog based on how far objects can be seen. Real-world datasets typically lack explicit visibility annotations, making them unsuitable for controlled evaluation. Using visibility levels allows for a systematic analysis of model performance under varying fog conditions and helps identify the point at which perception begins to degrade. To simulate varying visibility conditions, we augment the RGB images and lidar point clouds with three levels of synthetic fog, corresponding to three visibility distances of 30m, 50m, and 75m. 




To perform fog augmentation in point clouds, we employ the method by Hahner et al. \cite{hahner2021fogsim}. For the image augmentation, a custom approach was tailored based on \cite{duthon2019light}, \cite{sakar2018syntheticdata}, \cite{bermuth2019snowfog}. The method models fog as a population of droplets whose radii are sampled from a lognormal distribution, capturing the natural variability in particle sizes. For each droplet radius, wavelength-dependent extinction efficiencies are computed using Mie theory, which accounts for both scattering and absorption by spherical particles. These per-droplet extinction values are then multiplied by the droplet concentration to obtain the overall extinction coefficient of the fog medium. The depth maps corresponding to KITTI images were taken from \cite{tremblay2021rain}. This depth data is used to scale light attenuation based on the distance light travels through the fog. Using the computed extinction coefficients and pixel-wise distances, Beer–Lambert’s law is applied to determine the final pixel intensities. As a result, nearby objects appear less obscured, while distant objects are increasingly attenuated, producing realistic, depth-dependent fog effects. 



\subsection{Log-Likelihood Ratio construction }

\begin{figure}[H]
\centering
\includegraphics[width=0.9\textwidth]{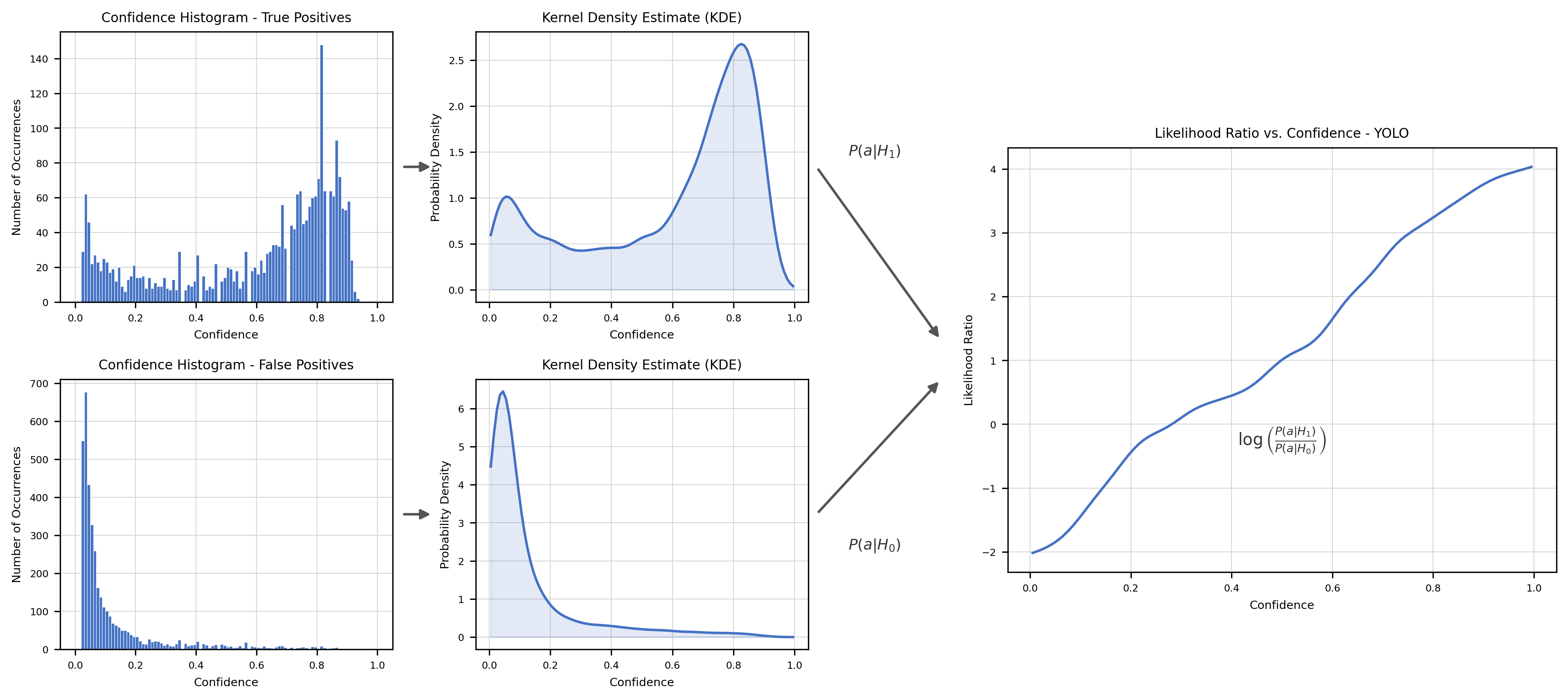}
\caption{LLR construction for YOLO}
\label{Fig_8}
\end{figure}

As described in section \ref{bayesian_fusion_impl}, inference is run on the KITTI clear weather training subset, comprising 4488 samples, using state-of-the-art pre-trained object detectors run at very high sensitivity. For each sensor, detection scores obtained after non-max suppression in empty areas (i.e., parts of the training images that do not contain ground truth objects) count towards the $H_0$ histogram of that sensor, while scores obtained for detections that have an intersection-over-union (IoU) greater than 0.5 with a ground truth bounding box count towards the $H_1$ histogram. Sample LLR construction for the camera detector is shown in Figure \ref{Fig_8}.

The histograms are converted into estimated probability density functions $p(a|H)$ using Kernel Density Estimation (KDE), with Scott’s rule determining the (locally adaptive) bandwidth of the kernel used to smooth the histogram. The logarithm of $p(a|H_1) / p(a|H_0)$ is then taken to obtain a single-sensor LLR for any value of confidence score $a$.




\subsection{Context Network Training}
The context extraction and adaptation networks are jointly trained on a mixture of clear and foggy data as discussed in section \ref{context_impl}. The training set comprises four variations of the original 4488 clear weather training samples—covering clear conditions and synthetic fog at visibility levels of 30m, 50m, and 75m for both images and point clouds, resulting in a total of 17952 samples. As mentioned earlier, model training is performed at the level of fused sensor detections, not individual frames. For each of the 4488 frames, detection outputs from YOLOv8 and PointPillars are associated to form fused detections. The resulting set includes independent class predictions, bounding boxes, confidence scores, and LLRs from each detector, along with fused bounding boxes, IoU, and is categorized into true positives and false positives by comparing to KITTI ground truth labels. The networks are trained with samples drawn from both true positive and false positive categories. Thus, the effective training set consists of 17952*n detection samples, where n is the average number of fused detections per frame.

The context network is trained using an end-to-end joint optimization approach. Based on the TP/FP distinction on the fused dataset, separate camera and lidar losses are computed using the loss function from \cite{moustakides2019lossfn} and summed to form a total loss. During the backward pass, gradients from the camera loss flow exclusively through the camera FFN, while gradients from the lidar loss flow exclusively through the lidar FFN. Both gradient pathways then converge at the shared ResNet-based context encoder, updating the weights across all components in a single, unified optimization step. For each training epoch, the model is validated using the validation subset consisting of clear and foggy dataset. The model is trained for 50 epochs using Stochastic Gradient Descent (SGD) as the optimizer, with a learning rate of $1\times 10^{-6}$. The trained context network is then integrated into the overall fusion architecture, and the complete model is evaluated using the test set. Each of the 1497 samples in the test set is presented under four variations of clear and foggy conditions, similar to the training set, resulting in a total of 5988 test instances.



\section{Results}
This research focuses on robust pedestrian detection under foggy environmental conditions using multimodal sensor data. We evaluate pedestrian detection performance across three configurations: 


\begin{itemize}[noitemsep, topsep=2pt]
    \item[\textbf{C1}] Individual object detectors operating independently.
    \item[\textbf{C2}] Fusion of independent object detectors using non-adaptive Bayesian fusion.
    \item[\textbf{C3}] Fusion of independent object detectors using context-adaptive Bayesian fusion.
\end{itemize}

This comparison highlights the impact of sensor fusion strategies from baseline performance of standalone detectors to improvements with non-adaptive fusion, and further gains achieved through context-aware adaptive fusion. Precision–Recall (PR) curves are used to quantify and visualize the comparative performance of the three candidate configurations. Area under the precision-recall curve is denoted as Average Precision (AP) and used as an indicator of general performance.

\begin{figure}[b]
\centering
\includegraphics[width=0.45\textwidth]{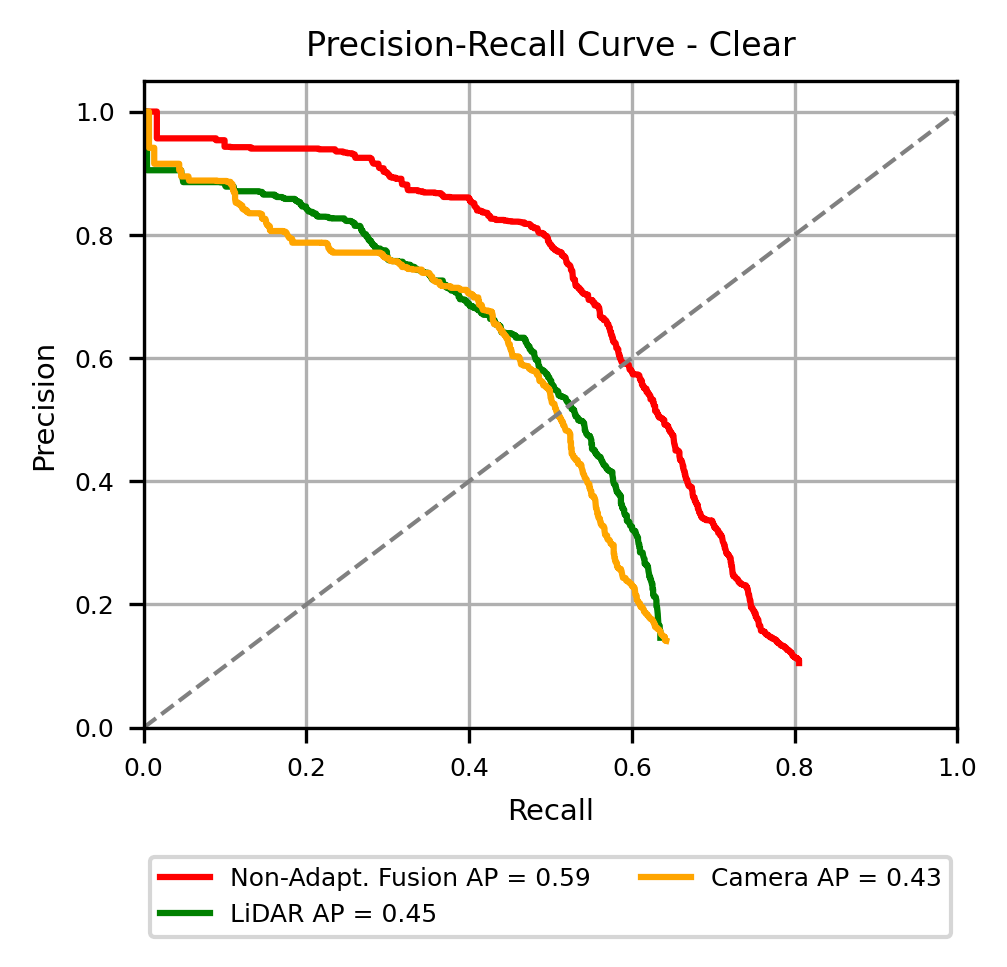}
\caption{Performance of pre-trained individual detectors and non-adaptive sensor fusion under clear weather conditions}
\label{Fig_10}
\end{figure}

The study in \cite{willems2024contextcompensation} compares the pedestrian detection performance of state-of-the-art image based object detectors with and without the calibration layer. In this study, we focus on multimodal sensor fusion. First, we compare the performance of pedestrian detection between the individual detectors (i.e., \textbf{C1}) and non-adaptive Bayesian late fusion (i.e., \textbf{C2}) pipeline under clear weather conditions (Figure \ref{Fig_10}). This fusion approach realizes significantly higher average precision than any single detector (fused AP of 0.59 compared to 0.45 for the best single-sensor detector), validating Bayesian fusion as an effective method to exploit complementarity in heterogeneous perception systems in non-challenging conditions.

\begin{figure}[t]
    \centering
    \subfigure[PointPillars]{\includegraphics[width=0.45\textwidth]{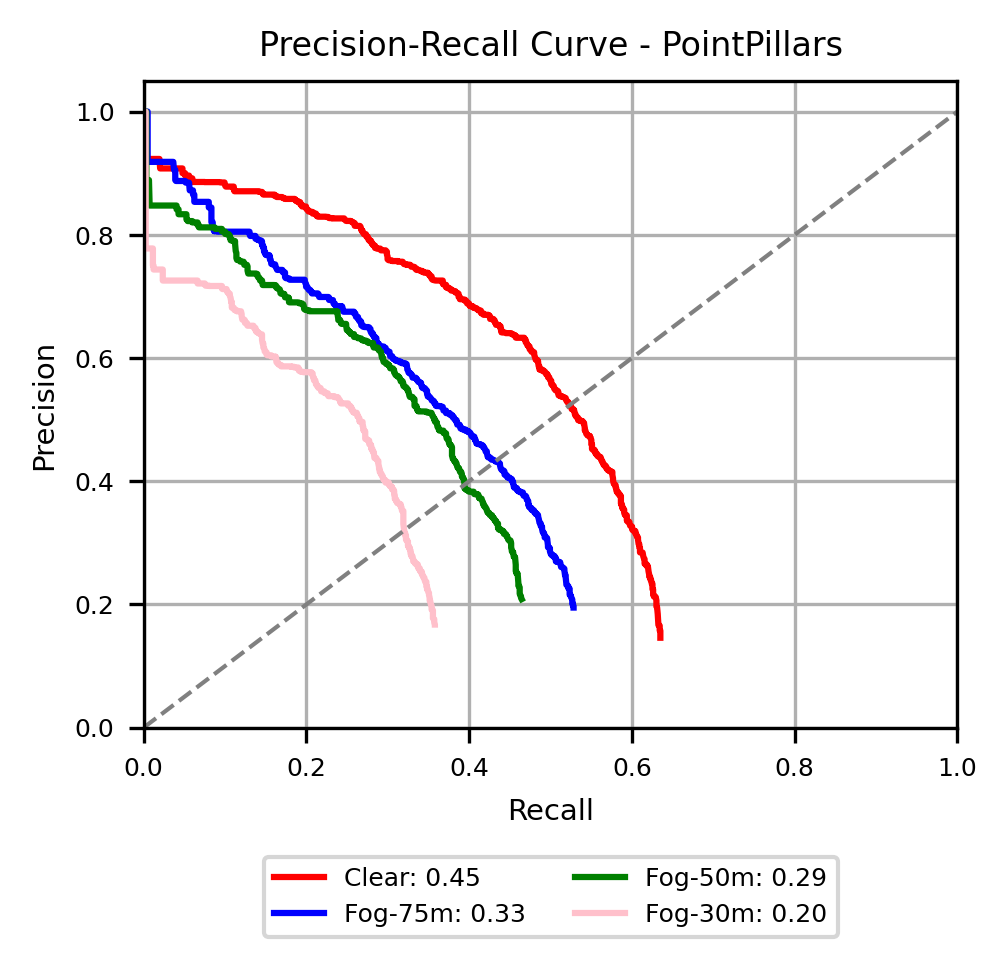}}
    \hfill
    \subfigure[YOLO]{\includegraphics[width=0.45\textwidth]{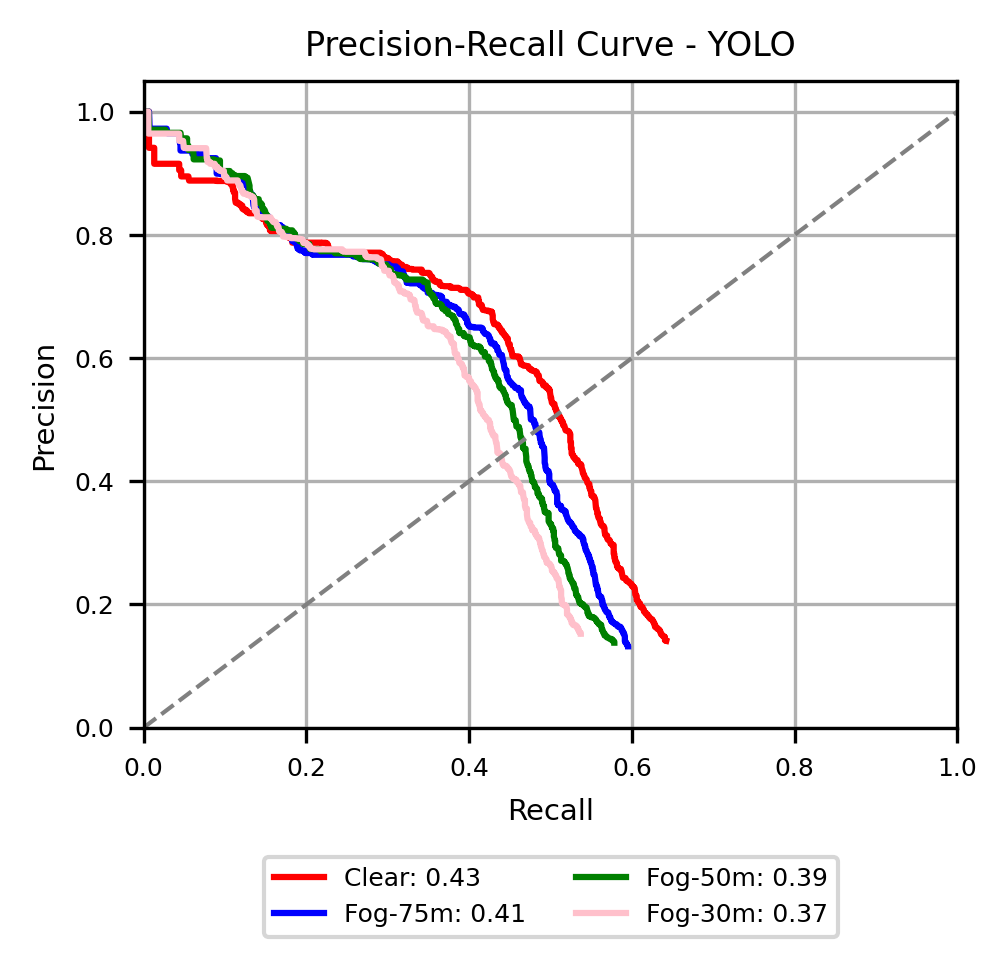}}
    \caption{Performance of pre-trained individual detectors on clear and foggy data in terms of AP}
    \label{Fig_14}
\end{figure}

\begin{figure}[htbp]
    \centering
    \subfigure[Clear]{\includegraphics[width=0.43\textwidth]{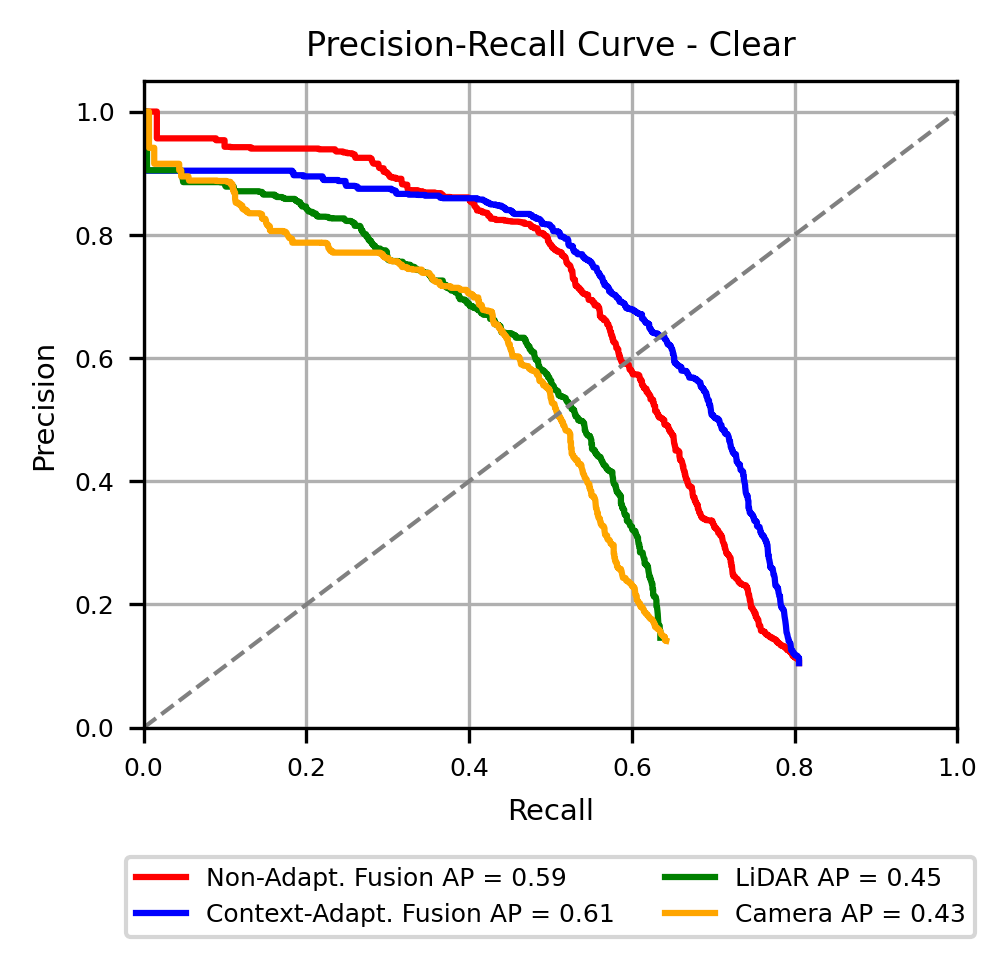}}
    \hfill
    \subfigure[75m Fog]{\includegraphics[width=0.43\textwidth]{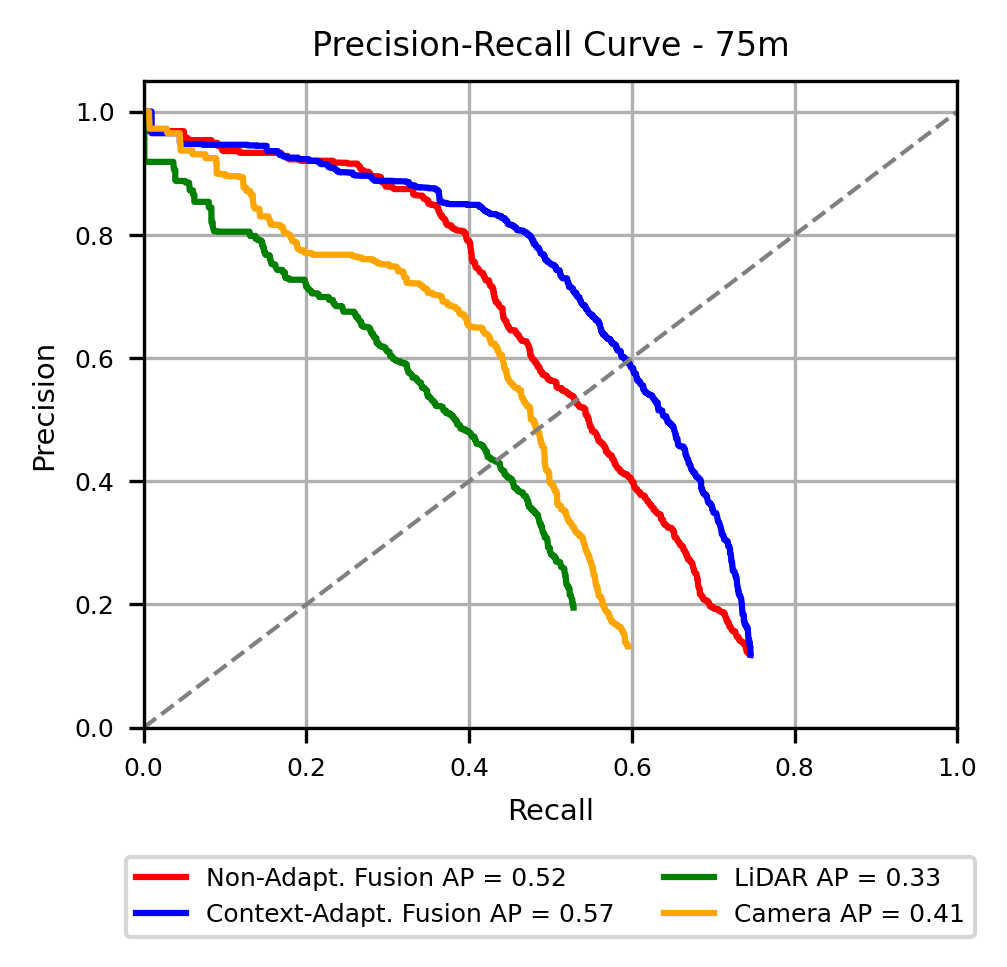}}
    
    \vspace{0.1cm}
    
    \subfigure[50m Fog]{\includegraphics[width=0.43\textwidth]{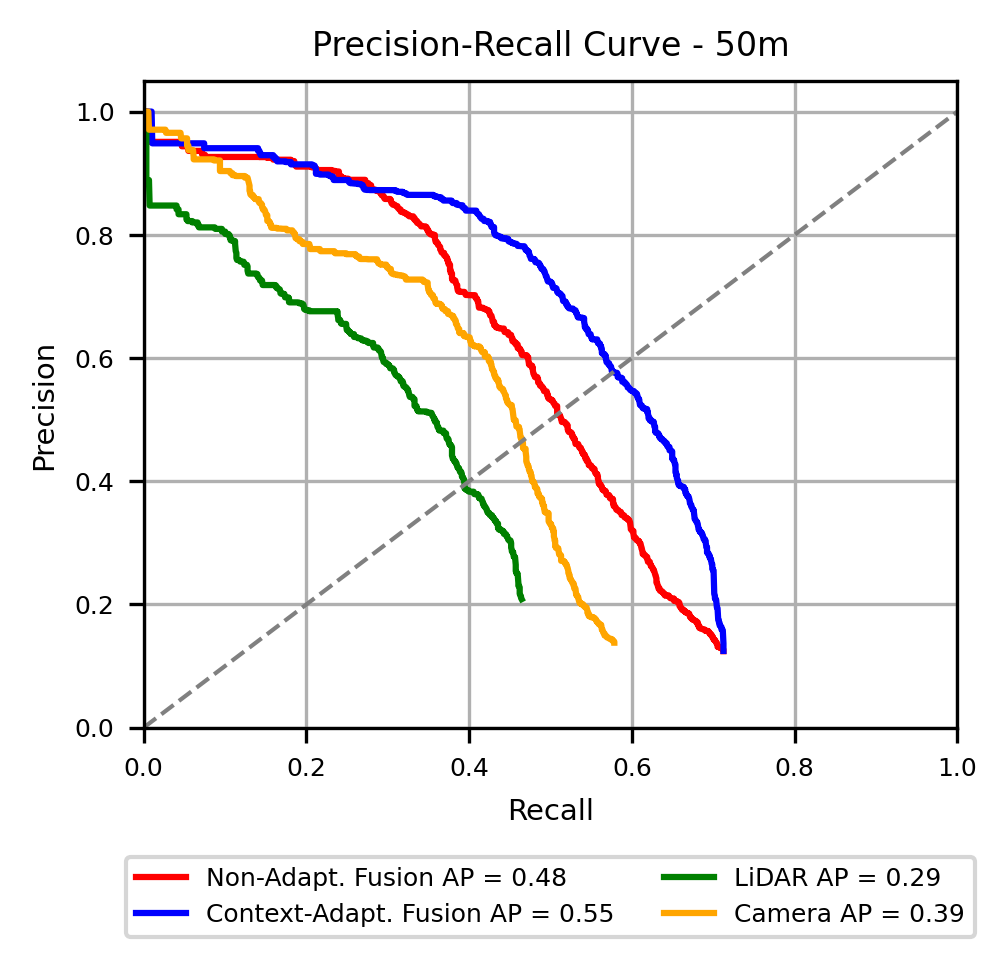}}
    \hfill
    \subfigure[30m Fog]{\includegraphics[width=0.43\textwidth]{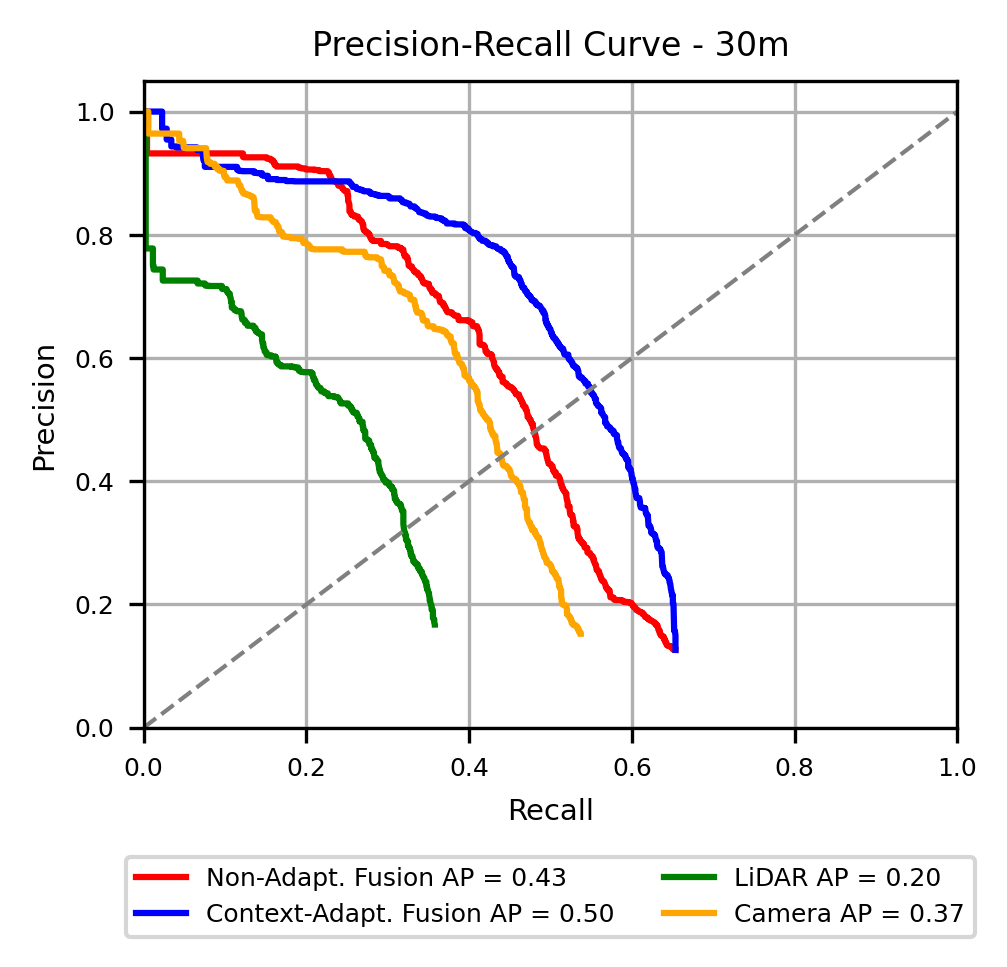}}

    \vspace{0.1cm}

    \subfigure[Combined]{\includegraphics[width=0.45\textwidth]{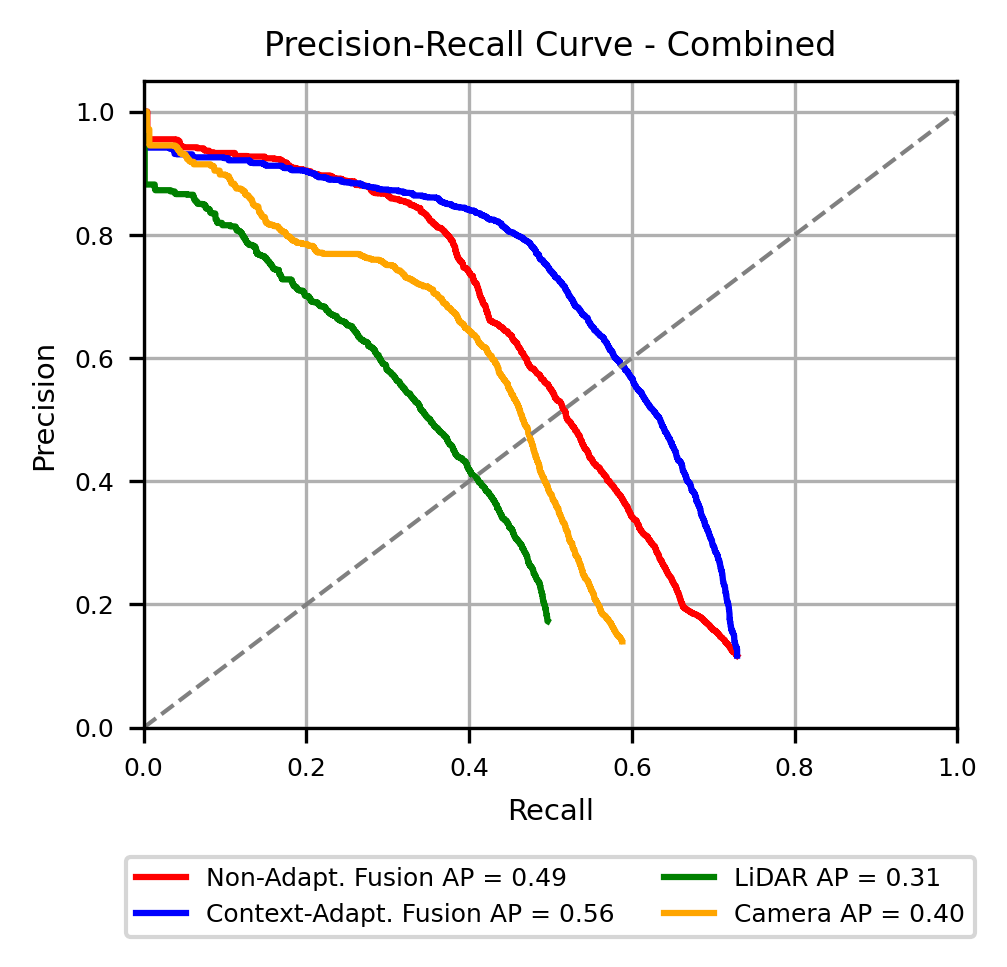}}

    \caption{PR curves comparing the performance of individual detectors, non-adaptive fusion, and context-aware fusion across (a) clear (b) 75m fog (c) 50m fog (d) 30m fog (e) Combined (all fog levels and clear)}
    \label{Fig_11}
    
\end{figure}

Figure \ref{Fig_14} shows the impact of (simulated) fog on the detection performance of PointPillars and YOLO (pre-trained only on clear-weather data). Both experience significant degradation on foggy data, with PointPillars most affected due to reduced point cloud density and reflectivity. Consequently, non-adaptive fusion also suffers from these degradations of the RGB camera and lidar modalities (Figure \ref{Fig_11}). 


In comparison, context-adaptive fusion with our proposed network (i.e., \textbf{C3}) maintains consistent performance. While non-adaptive fusion still improves performance compared to individual detectors across different fog levels, context-adaptive fusion demonstrates a significantly larger advantage. Our context-adaptive approach achieves an average precision (AP) of $\sim0.56$ across all fog densities, and offers a $\sim14\%$ relative improvement in average precision over non-adaptive fusion, showing the importance of context for sensor fusion (Table \ref{tab:sensor_fusion_ap}). The context-adaptive model was trained on both clear and multi-level fog conditions, enabling it to perform similar to non-adaptive fusion in clear weather and outperform it under foggy conditions. This highlights that a single context-aware model can effectively handle both clear and foggy scenarios.



\begin{figure}[H]
\centerline{\includegraphics[width=0.7\columnwidth]{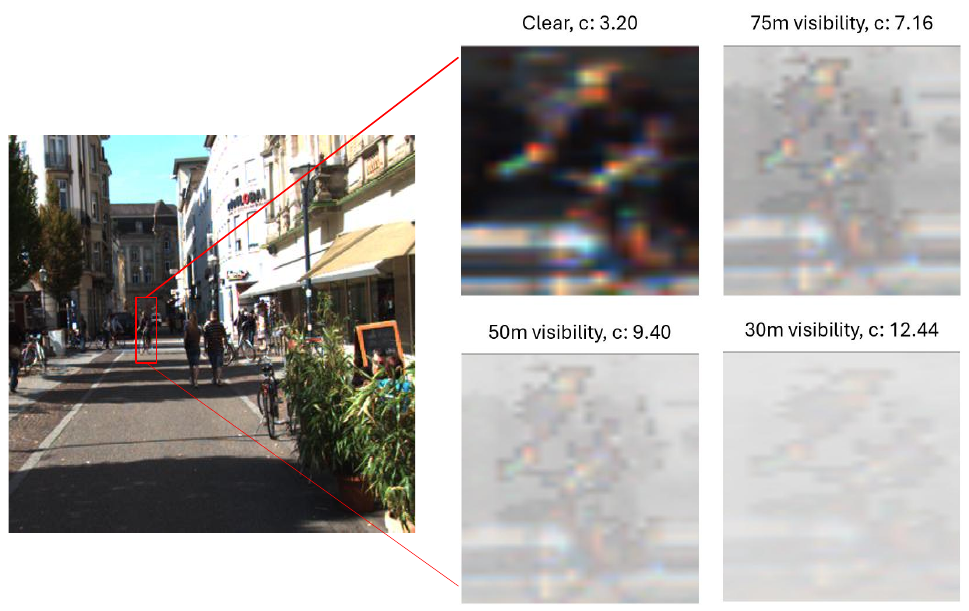}}
\caption{Variation of context value under different conditions for a cyclist.}
\label{Fig_13}
\end{figure}

Another key aspect is the role of the context extraction network (i.e., ResNet-18) in influencing modality-specific weights. Figure \ref{Fig_13} illustrates how the context value varies with environmental conditions. For example, in the case of a cyclist, the context value increases as the detection task becomes more challenging. As fog density rises—from clear conditions to a lower visibility level of 30m (i.e., dense fog)—the context value grows accordingly, reflecting the network’s ability to capture scene difficulty.

\begin{table}[t!]
\centering
\begin{tabular}{lcccc}
\toprule
 & \textbf{\begin{tabular}[c]{@{}c@{}}Context-Adaptive\\ Fusion (AP)\end{tabular}} 
 & \textbf{\begin{tabular}[c]{@{}c@{}}Non-adaptive\\ Fusion (AP)\end{tabular}} 
 & \textbf{\begin{tabular}[c]{@{}c@{}}Camera\\ (AP)\end{tabular}} 
 & \textbf{\begin{tabular}[c]{@{}c@{}}Lidar\\ (AP)\end{tabular}} \\ 
\midrule
Clear    & 0.61 & 0.59 & 0.43 & 0.45 \\
75m fog  & 0.57 & 0.52 & 0.41 & 0.33 \\
50m fog  & 0.55 & 0.48 & 0.39 & 0.29 \\
30m fog  & 0.50 & 0.43 & 0.37 & 0.20 \\
Combined & 0.56 & 0.49 & 0.40 & 0.31 \\
\bottomrule
\end{tabular}
\caption{Comparison of Average Precision (AP) across different visibility conditions}
\label{tab:sensor_fusion_ap}
\end{table}

\section{Conclusion}

This research addresses the challenge of robust pedestrian detection under foggy conditions using multimodal sensor data. Our findings demonstrate that conventional non-adaptive sensor fusion fails to deliver significant improvements when individual detectors degrade under foggy conditions. To overcome this limitation, we introduced a context-adaptive Bayesian late fusion framework that extracts environmental context and dynamically adjusts modality-specific weights to comply with diverse conditions. The model effectively generalizes across multiple visibility conditions with a limited dataset, learning features for varying visibility without requiring manual annotations of weather or visibility. By decoupling context modeling from the detection pipeline, our approach enables adaptation to different contexts and scenarios while avoiding the significant cost and difficulty of re-training detectors for different contexts.  Experimental results confirm that context-adaptive fusion outperforms both individual detectors and non-adaptive fusion, providing an elegant solution to the challenge of strong variation of conditions in which real-world autonomous perception systems must operate. Future work will cover deeper analysis of the extracted context as well as testing real-world adverse weather scenarios.

\bibliography{bmvc}
\end{document}